\documentclass{article}
\usepackage{microtype}
\usepackage{graphicx}
\usepackage{subfigure}
\usepackage{booktabs} 
\usepackage{color}
\usepackage{amsmath,amssymb,amsthm,mathtools}
\usepackage{natbib}
\newtheorem{theorem}{\textit{Theorem}}
\newtheorem{definition}{\textit{Definiton}}
\newtheorem{lemma}[theorem]{\textit{Lemma}}
\newtheorem{assumption}{\textit{Assumption}}
\newtheorem{proposition}[theorem]{\textit{Proposition}}
\DeclareMathOperator*{\argmin}{arg~min}

\definecolor{customgray}{rgb}{0.25,0.25,0.25}
\definecolor{customred}{rgb}{0.8,0.05,0.05}
\newcommand{\best}[1]{\textcolor{customred}{\textbf{#1}}}
\newcommand{\std}[2]{\large{#1} \textcolor{customgray}{\normalsize{$\pm$#2}}}

\newcommand{\Eqref}[1]{Eq.~(\ref{#1})}
\newcommand{\Thmref}[1]{Theorem~\ref{#1}}
\newcommand{\Propref}[1]{Proposition~\ref{#1}}
\newcommand{\Lemref}[1]{Lemma~\ref{#1}}

\newcommand{\Tabref}[1]{Table~\ref{#1}}
\newcommand{\mE}{\mathbb{E}}
\newcommand{\mV}{\mathbb{V}}
\newcommand{\mP}{\mathbb{P}}
\newcommand{\indep}{\mathop{\perp\!\!\!\!\perp}}

\newcommand{\trueR}{\mathcal{R}_{true}}
\newcommand{\estR}{\widehat{\mathcal{R}}}
\newcommand{\ipm}{\mathrm{IPM}_G}
\newcommand{\hattau}{\hat{\tau}}
\newcommand{\tiltau}{\tilde{\tau}}
\newcommand{\obsY}{Y}
\newcommand{\oneY}{Y(1)}
\newcommand{\zeroY}{Y(0)}
\newcommand{\tY}{Y(t)}
\newcommand{\oneM}{m_1(X)}
\newcommand{\zeroM}{m_0(X)}
\newcommand{\oneW}{w_1(X)}
\newcommand{\zeroW}{w_0(X)}
\newcommand{\oneP}{p_1(x)}
\newcommand{\zeroP}{p_0(x)}
\newcommand{\tP}{p_t(x)}
\newcommand{\cftP}{p_{1-t}(x)}
\newcommand{\tPrep}{p_{t}^{\Phi}}
\newcommand{\cftPrep}{p_{1-t}^{\Phi}}

\newcommand{\tM}{m_t(x)}
\newcommand{\oneF}{f_1(X)}
\newcommand{\zeroF}{f_0(X)}
\newcommand{\tF}{f_t(x)}
\newcommand{\xty}{X,T,Y}
\newcommand{\ps}{e(X)}

\newcommand{\tELL}{\ell_{h, \Phi}^w}
\newcommand{\fEPS}{\epsilon^w_{F} (h, \Phi)}
\newcommand{\cfEPS}{\epsilon^w_{CF} (h, \Phi)}
\newcommand{\foneEPS}{\epsilon^w_{F_1} (h, \Phi)}
\newcommand{\cfoneEPS}{\epsilon^w_{CF_1} (h, \Phi)}
\newcommand{\fzeroEPS}{\epsilon^w_{F_0} (h, \Phi)}
\newcommand{\cfzeroEPS}{\epsilon^w_{CF_0} (h, \Phi)}
\newcommand{\ftEPS}{\epsilon^w_{F_t} (h, \Phi)}

\newcommand{\Brep}{B_{\Phi}}
\newcommand{\givenX}{\ | \ X}
\newcommand{\mC}{C_{\mathrm{max}}}

\usepackage{color}

\usepackage{hyperref}

\usepackage[accepted]{icml2020}

\icmltitlerunning{Counterfactual Cross-Validation}

\begin{document}

\twocolumn[
\icmltitle{Counterfactual Cross-Validation: \\
Stable Model Selection Procedure for Causal Inference Models}


\icmlsetsymbol{equal}{*}
\begin{icmlauthorlist}
\icmlauthor{Yuta Saito}{titech}
\icmlauthor{Shota Yasui}{ca}
\end{icmlauthorlist}

\icmlaffiliation{titech}{Tokyo Institute of Technology,}
\icmlaffiliation{ca}{CyberAgent, Inc}
\icmlcorrespondingauthor{Yuta Saito}{saito.y.bj@m.titech.ac.jp}
\icmlcorrespondingauthor{Shota Yasui}{yasui\_shota@cyberagent.co.jp}

\icmlkeywords{CATE Prediction, Causal Inference, Model Selection, Hyperparameter tuning}

\vskip 0.3in
]



\printAffiliationsAndNotice{}  

\begin{abstract}
We study the model selection problem in \textit{conditional average treatment effect} (CATE) prediction. Unlike previous works on this topic, we focus on preserving the rank order of the performance of candidate CATE predictors to enable accurate and stable model selection. To this end, we analyze the model performance ranking problem and formulate guidelines to obtain a better evaluation metric. We then propose a novel metric that can identify the ranking of the performance of CATE predictors with high confidence. Empirical evaluations demonstrate that our metric outperforms existing metrics in both model selection and hyperparameter tuning tasks.
\end{abstract}

\section{Introduction}

Predicting conditional average treatment effect (CATE) for certain actions is essential for optimizing metrics of interest in various domains. 
In digital marketing, incrementality is becoming increasingly important as a performance metric~\cite{diemert2018large}. 
For instance, for a given product, users who will be shown its ads should be chosen based on CATE. It can help avoid showing ads to a user who will buy that product even without seeing the ads. 
There can be significant applications of CATE prediction in the healthcare segment as well~\cite{alaa2017bayesian}. 
This is because, for pursuing an optimal precision medicine, we need to know which treatments will be beneficial or harmful for a particular patient.

To achieve high-accuracy CATE prediction, one has to address the fundamental problem of causal inference, which is that both treated and untreated outcomes can never be observed simultaneously from the same unit~\cite{holland1986statistics}. 
Hence we are unable to observe a causal effect and to use it as label to train prediction models. 
Most previous studies related to the CATE prediction focused on developing methods that can address this fundamental problem and achieve high prediction accuracy~\cite{yoon2018ganite,yao2018representation,louizos2017causal,shalit2017estimating,du2019adversarial,alaa2018limits}. 

In model evaluation and selection, the fundamental problem of causal inference poses an additional critical challenge. 
Because labels are not observed directly, we are unable to calculate loss metrics such as \textit{mean squared error} (MSE). 
Therefore, data-driven validation procedures such as cross-validation are not directly applicable to model selection and hyperparameter tuning of CATE prediction models. 
This makes it challenging to identify the suitable model and appropriate hyperparameter values that should be used when applying CATE prediction to real-world problems.

Several prior studies tackle the model evaluation problem in CATE prediction.
~\cite{gutierrez2017causal} proposed using the inverse probability weighting (IPW) outcome as the pseudo-label for the true CATE for the calculation of an evaluation metric.
~\cite{schuler2018comparison} used the loss function of R-learner~\cite{nie2017quasi} for the evaluation.
~\cite{alaa2019validating} used influence functions to obtain a more efficient estimator for the loss. 
These works are mainly focused on improving the accuracy of estimating the evaluation metric of interest. 

Unlike previous works, we focus on choosing the best model or hyperparameters from potential candidates. 
For this purpose, \textbf{\textit{we only need to know the rank order of the performance of candidate predictors}}, which is easier than directly estimating the true performance.
To achieve this, we first theoretically analyze the problem of ranking the true performance of CATE predictors and identify the conditions that an ideal metric should satisfy. 
Building on the analysis, we propose a novel evaluation procedure that preserves the true performance ranking of candidate predictors and minimizes the upper bound of the finite sample uncertainty in model selection.
Through empirical evaluations, we demonstrate that the proposed metric performs better than existing heuristic metrics in model selection and hyperparameter tuning of CATE predictors.

\section{Related Work}
CATE prediction has been extensively studied by combining causal inference and machine learning techniques aiming for the best possible personalization of interventions. 
State-of-the-art approaches are constructed by utilizing the adversarial generative model, Gaussian process, deep neural networks, and latent variable models~\cite{yoon2018ganite,alaa2018limits,louizos2017causal,alaa2017bayesian,hassanpour2019counterfactual,hassanpour2020Learning,shi2019adapting,bica2020estimating,yao2020survey}. 
Among the diverse methods that predict CATE from observational data, the approach that is most related to this work is the method based on representation learning~\cite{bengio2013representation,johansson2020generalization}. 
All methods based on representation learning attempt to map the original feature vectors into the desirable latent representation space so that it eliminates selection biases. 
Balancing neural network~\cite{johansson2016learning} is the most basic method that uses discrepancy distance~\cite{mansour2009domain}, a domain discrepancy measure in unsupervised domain adaptation for the regularization term. 
Counterfactual regression~\cite{shalit2017estimating} minimizes the upper bound of the ground-truth loss for the CATE by utilizing an integral probability metric~\cite{sriperumbudur2012empirical}. 
In addition to these, methods that obtain a latent representation by preserving a pairwise similarity~\cite{yao2018representation,yao2019ace} or by applying adversarial learning~\cite{du2019adversarial} have been proposed.  

The prediction methods stated above have provided promising results on standard benchmark datasets. 
However, previous studies have evaluated such CATE predictors by using synthetic datasets or simple heuristic metrics such as policy risk ~\cite{yoon2018ganite,shalit2017estimating,yao2018representation}. 
However, these evaluations do not give a definitive answer about which models would actually be best suited for a given real-world dataset~\cite{alaa2019validating,setoguchi2008evaluating}. 
Therefore, to bridge the gap between CATE prediction and applications, developing a reliable evaluation metric is critical. 

There are only a few studies directly tackling the evaluation problem of CATE prediction models.
\cite{schuler2018comparison} conducted an extensive survey of several heuristic metrics and provided experimental comparisons. 
In particular, they introduced inverse probability weighting (IPW) validation, which utilizes an unbiased estimator for the true CATE as an alternative to the true causal effects, and $\tau$-risk, which is based on a loss function of R-learner~\cite{nie2017quasi}. 
In addition, they showed that these metrics empirically outperformed another naive metric, $\mu$-risk, which estimates predictive risk separately for treated and control outcomes using only factual samples. 
In contrast,~\cite{rolling2014model} proposed a propensity matching-based metric called TECV and showed its consistency to the true ranking of the performance of CATE prediction models. 
However, they did not analyze the uncertainty of the metric, such as its asymptotic variance.
It was also empirically outperformed by IPW validation~\cite{schuler2018comparison}. 
Nonetheless,~\cite{alaa2019validating} improved heuristic plug-in metrics by introducing a meta-estimation technique using influence functions in a theoretically sophisticated manner. 
Our proposed metric can be further improved by an estimation method based on influence functions. 

All the existing metrics aim to estimate the true metric of interest directly, or they do not consider the uncertainty in model selection. 
However, to conduct accurate model selection and hyperparameter tuning, it is essential to rank model performance accurately, although the aforementioned metrics do not always guarantee the preservation of such rankings. 
Moreover, analysis of the uncertainty of the model evaluation is necessary, especially in domains in which the size of the validation datasets might be small (e.g., education or public health). 
Therefore, in contrast to previous works, we investigate a method to accurately preserve the rank order of performance of the candidate predictors while also analyzing the finite sample uncertainty in model selection.

\section{Setup}
We denote $X \in \mathcal{X} \subseteq \mathbb{R}^{d}$ as a $d$-dimensional feature vector and $T \in \mathcal{T} = \{0, 1\}$ as a binary treatment assignment indicator. 
When an individual $i$ receives treatment, then $T_i = 1$, otherwise, $T_i = 0$. 
We follow the \textit{potential outcome framework}~\cite{rosenbaum1983central,rubin2005causal,imbens2015causal} and assume that there exist two potential outcomes denoted as $ \zeroY, \oneY \in \mathcal{Y} \subseteq \mathbb{R} $ for each individual. 
$\zeroY$ is a potential outcome associated with $T=0$, and $\oneY$ is associated with $T=1$. 
Note that each individual receives only one treatment and reveals the outcome value for the received treatment. 
We use $p (X, T, \zeroY, \oneY )$, or simply $p$, to denote the joint probability distribution of these random variables.

We formally define the \textit{conditional average treatment effect} (CATE) for a given feature vector $x \in \mathcal{X}$ as:
\begin{align*}
  \tau ( x ) \coloneqq \mE [ \oneY - \zeroY  \ | \ X = x ].
\end{align*}
In addition, we use some notations to represent parameters of $p$.
First, we define the expected potential outcomes conditioned on a feature vector $x \in \mathcal{X}$ as:
\begin{align*}
  \tM \coloneqq  \mE_{\tY} [ \tY \ | \ X = x ], \; \forall t \in \{0, 1\}.
\end{align*}
Next, we define the \textit{propensity score} as:
\begin{align*}
  e ( x ) \coloneqq  \mP \left( T = 1 \ | \ X = x \right).
\end{align*}
This parameter is widely used to estimate treatment effects from observational data~\cite{rosenbaum1983central,rubin1974estimating,imbens2015causal}.

Throughout the paper, we make the following standard assumptions in causal inference:
\begin{assumption}(Unconfoundedness)
Potential outcomes $( \zeroY, \oneY )$ are independent of the treatment assignment indicator $T$ conditioned on the feature vector $X$, i.e.,
\begin{align*}
 \zeroY, \oneY  \indep T \ | \ X  .
\end{align*} 
\end{assumption}

\begin{assumption}(Overlap)
For any $x \in \mathcal{X}$, the true propensity score is strictly between 0 and 1, i.e., $0 < e (x) < 1  $.
\end{assumption}

\begin{assumption}(Consistency) 
Observed outcome $Y$ is represented using the potential outcomes and treatment assignment indicator as follows:
\begin{align*}
    Y = T \oneY + (1 - T) \zeroY  .
\end{align*}
\end{assumption}

Under these assumptions, the CATE is identifiable from observational data,
i.e., $ \tau(x) = \mE [ Y \, |\, X=x, T=1 ] - \mE [ Y \, |\, X=x, T=0 ] $.

Furthermore, we define some essential notations following~\cite{shalit2017estimating}.

\begin{definition} (Representation Function) 
$ \Phi : \mathcal{X} \rightarrow \mathcal{R} $ is a representation function and $\mathcal{R}$ is called the representation space. 
We assume that $\Phi $ is a twice differentiable, one-to-one function. 
Moreover, $\tPrep \coloneqq p (r | t=1)$ and $\cftPrep \coloneq p (r | t=0)$ are feature distributions for the treated and for the controlled induced over the representation space.
We also have $\Psi : \mathcal{R} \rightarrow \mathcal{X} $ as the inverse of $\Phi$, where $\Psi(\Phi(x))=x, \forall x \in \mathcal{X}$.
\end{definition}

\begin{definition} (Factual and Counterfactual Loss Functions) 
Let $h: \mathcal{R} \times \mathcal{T} \rightarrow \mathcal{Y}$ be a hypothesis, $w: \mathcal{X} \rightarrow \mathbb{R}_{\ge 0}$ be a weighting function, and $L: \mathcal{Y} \times \mathcal{Y} \rightarrow \mathbb{R}_{\ge 0}$ be a loss function. 
In addition, the expected loss for the unit and treatment pair $(x, t) \in \mathcal{X} \times \mathcal{T}$ is denoted as: 
\begin{align*}
    \tELL (x, t) \coloneqq \mE_{\tY} \left[ w(x) L (\tY, h(\Phi(x), t) ) \givenX = x \right] .
\end{align*}
where we use the squared loss: $L(y,y^{\prime}) = (y-y^{\prime})^2$, hereinafter.
Then, the expected factual and counterfactual losses of a combination of a hypothesis $h$ and a representation function $\Phi$ are defined as:
\begin{align*}
    \fEPS & \coloneqq  \int_{\mathcal{X} \times \mathcal{T}} \tELL(x, t) p(x, t) d x d t, \\
    \cfEPS & \coloneqq  \int_{\mathcal{X} \times \mathcal{T}} \tELL(x, t) p(x, 1-t) d x d t .
\end{align*}
where $F$ and $CF$ stand for factual and counterfactual, respectively.
Further, the expected factual and counterfactual losses on the treated ($t=1$) and on the controlled ($t=0$) are represented as:
\begin{align*}
    \foneEPS & \coloneqq  \int_{\mathcal{X}} \tELL(x, t=1) \oneP dx , \\
    \fzeroEPS & \coloneqq  \int_{\mathcal{X}} \tELL(x, t=0) \zeroP dx , \\
    \cfoneEPS & \coloneqq \int_{\mathcal{X}} \tELL(x, t=0) \oneP dx , \\
    \cfzeroEPS & \coloneqq  \int_{\mathcal{X}} \tELL(x, t=1) \zeroP dx .
\end{align*}
where $\tP \coloneqq  p (x \ | \ T=t) $.

By the definition of the conditional probability, the following equations hold for factual and counterfactual losses:
\begin{align*}
    \fEPS & = \pi_1 \cdot \foneEPS + \pi_0 \cdot \fzeroEPS , \\
    \cfEPS & = \pi_1 \cdot \cfoneEPS + \pi_0 \cdot \cfzeroEPS ,
\end{align*}
where $\pi_t \coloneqq  \mP (T = t)$.
\end{definition}

We also define a class of metrics between probability distributions~\cite{sriperumbudur2012empirical}.
\begin{definition}(Integral Probability Metric)
For two probability density functions defined over a space $\mathcal{S} \subseteq \mathbb{R}^d$ and for a family of functions $G \coloneqq \{g : \mathcal{S} \rightarrow \mathbb{R} \}$, the IPM between the two density functions $p$ and $q$ is defined as:
\begin{align*}
    \ipm (p, q ) \coloneqq  \sup _{g \in G}\left|\int_{\mathcal{S}} g(s)\left(p(s)-q(s)\right) d s\right| .
\end{align*}
Function families $G$ can be the family of bounded continuous functions, the family of 1-Lipschitz functions, and the unit-ball of functions in a universal reproducing Hilbert kernel space.
\end{definition}

\subsection{Evaluation of CATE prediction models}
In previous studies~\cite{gutierrez2017causal,schuler2018comparison,alaa2019validating}, the evaluation of a CATE predictor $ \hattau (\cdot)  $ has been formulated as accurately estimating the following ground-truth performance metric from a size $n$ of i.i.d observational validation dataset $\mathcal{V} = \{X_i, T_i, Y_i\}_{i=1}^{n}$:
\begin{align}
  \trueR ( \hattau ) 
  & \coloneqq  \mE_{X} \left[ L \left( \tau(X) , \hattau(X) \right) \right] \notag \\
  & = \mE_{X} \left[ \left( \tau(X) - \hattau(X) \right)^2 \right], \label{eq:true_metric}
\end{align}
where $\trueR ( \hattau )$ is the true performance metric of $ \hattau( \cdot ) $\footnote{\Eqref{eq:true_metric} is also termed as the \textit{expected precision in estimation of heterogeneous effect} (PEHE).}.

This approach is intuitive and ideal. 
However, realizations of the true CATE are never observable, and thus, accurate performance estimation is difficult. 
Moreover, estimating the true metric values is not always necessary to conduct valid model selection or hyperparameter tuning. 
It may be possible to obtain a better evaluation metric under an objective specific to selection and tuning. 
Thus, we take a different approach from previous works and aim to construct a performance estimator $ \estR \left( \hattau \right) $ satisfying the following condition:
\begin{align}
    \trueR \left( \hattau \right) \le \trueR \left( \hattau^{\prime} \right) \Rightarrow 
    \estR \left( \hattau \right) \le \estR \left( \hattau^{\prime} \right), \; \forall \,  \hattau, \hattau^{\prime} \in \mathcal{M}. \label{eq:preserve_ranking}
\end{align}
where $\mathcal{M} = \{\hattau_1, ..., \hattau_{| \mathcal{M}|} \}$ is a set of candidate CATE predictors.

An estimator satisfying \Eqref{eq:preserve_ranking} gives an accurate ranking of candidate predictors by the ground-truth metric, and we can identify the best model among $\mathcal{M}$ using such an estimator. 
Our goal is to construct a sophisticated method to obtain a performance estimator that can help achieve the condition described in \Eqref{eq:preserve_ranking} to enable accurate model selection of CATE predictors.
\section{Method}
To achieve our goal, we consider the following feasible estimator for the ground-truth performance metric:
\begin{align}
  \estR \left( \hattau \right)
  \coloneqq  \frac{1}{n} \sum_{i=1}^n  \left( \tiltau \left( X_i, T_i, Y_i \right) - \hattau \left( X_i \right) \right)^2
  \label{eq:performance_estimator}
\end{align}
where $ \tiltau \left( \cdot \right) $ is the \textit{plug-in tau} and is calculated by using validation set. 
We consider the estimator in the form of \Eqref{eq:performance_estimator}, because it can be applied to estimating the performance of a method directly predicting CATE such as R-learner~\cite{nie2017quasi} and doubly robust learner~\cite{foster2019orthogonal}. 

Under our formulation, we aim to answer the following research question: \textbf{\textit{What is the best plug-in tau to rank the performance of given candidate CATE predictors from an observational validation dataset?}}

To address this question, in Section~\ref{sec:what_is_the_good_plug-in_tau}, we theoretically analyze the performance estimator in the form of \Eqref{eq:performance_estimator} and identify the conditions that an ideal \textit{plug-in tau} should satisfy.
Then, in Section~\ref{sec:obtaining_plug-in_tau}, we propose a method to obtain a \textit{plug-in tau} that results in an accurate ranking of the true performance of candidate CATE predictors.

\subsection{What is the good \textit{plug-in tau}?} \label{sec:what_is_the_good_plug-in_tau}
First, the following proposition states that a \textit{plug-in tau} that is unbiased for the true CATE provides a desirable property of the resulting performance estimator.
\begin{proposition} Suppose that a given plug-in tau is an unbiased estimator for the true CATE  (i.e., $ \mE \left[ \tiltau \left( X, T, Y \right) \givenX \right] = \tau(X)$), then, the expectation of the performance estimator $ \estR $ is decomposed into the true performance metric and the MSE of the given plug-in tau:
\begin{align}
    \mE \left[  \estR \left( \hattau \right) \right]
    = \trueR \left( \hattau \right) + \underbrace{\mE \left[ \left( \tau(X) - \tiltau (X, T, Y) \right)^2 \right]}_{\text{independent of } \hattau} \label{eq:identify_rank}
\end{align}
See Appendix~\ref{sec:proof_rank_identification} for the proof.
\label{prop:rank_identification}
\end{proposition}

The first term of RHS of \Eqref{eq:identify_rank} is the true performance metric, and the second term is independent of the given predictor. Therefore, the expectations of the performance estimators preserve the difference between the true metric values as follows:
\begin{align*}
  \mE \left[  \estR \left( \hattau_1 \right) \right] - \mE \left[  \estR \left( \hattau_2 \right) \right]
  = \trueR \left( \hattau_1 \right) - \trueR \left( \hattau_2 \right) 
\end{align*}
where $\hattau_1  ,\, \hattau_2 \in \mathcal{M} $ are arbitrary candidate predictors.
This property is desirable, because the predictor that has the smallest expected value of $\estR$ among candidate predictors also has the smallest value of $\trueR$ among them; one can expect to select the best predictor among a set of candidates.

However, the expectation of the performance estimator is incalculable, because we can use only a finite sample validation set.
This motivates us to consider the finite sample uncertainty of the performance estimator. 
The empirical version of the performance estimator can be decomposed as
\begin{align}
  & \estR \left( \hattau \right) \notag \\
  & = \underbrace{\frac{1}{n} \sum_{i=1}^n (\tau(X_i)-\hattau(X_i))^2}_{\text{converges to } \trueR(\hattau)} \notag \\
  & \; - \underbrace{\frac{2}{n} \sum_{i=1}^{n} \left(\hattau \left(X_{i} \right) - \tau \left(X_{i} \right) \right) \left( \tiltau \left(X_{i}, T_i, Y_i \right)-\tau \left(X_{i} \right)\right) }_{\mathcal{W}: \text{source of uncertainty}} \notag \\
  & \; + \underbrace{ \frac{1}{n} \sum_{i=1}^n (\tau(X_i)-\tiltau(X_i, T_i, Y_i))^2}_{\text{independent of } \hattau}. \label{eq:estimator_decomposition}
\end{align}
In the RHS of \Eqref{eq:estimator_decomposition}, $\mathcal{W}$ is critical to the uncertainty and is controllable by $\tiltau$.
Thus, we try to minimize the variance of $\mathcal{W}$ with the aim of minimizing the uncertainty in model selection.
The following theorem upper bounds the variance of $\mathcal{W}$.
\begin{theorem}
Suppose that the plug-in tau is unbiased for the CATE and the output of the plug-in tau for an instance is independent of that of other instances. Then, we have the upper bound of the variance of $\mathcal{W}$ as follows:
  \begin{align}
    \mV\left( \mathcal{W} \right)  
    \le 4 \mC n^{-1} \  \mE_X \left[ \mV \left(\tiltau (X, T, Y) \givenX  \right) \right], \label{eq:variance_upper_bound}
  \end{align}
  where $ \mC = \sup_{x \in \mathcal{X}} ( \tau(x) - \hattau(x) )^2$.
  See Appendix~\ref{sec:proof_variance_upper_bound} for the proof.
\label{thm:variance_upper_bound}
\end{theorem}
In \Eqref{eq:variance_upper_bound}, the expected conditional variance of $\tiltau$ is controllable by the construction of the \textit{plug-in tau}.
Thus, a \textit{plug-in tau} satisfying the following condition is desirable to construct a stable performance estimator:
\begin{align}
   \min_{\tiltau \in \Theta}  \ & \mE_X \left[ \mV \left(\tiltau (X, T, Y) \givenX  \right) \right], \notag \\
  \text{s.t.} \ & \mE[\tiltau (\xty) \givenX ] = \tau(X). \label{eq:good_plug-in_tau}
\end{align}
where $\Theta$ is a pre-defined class of \textit{plag-in tau}.

A performance estimator using a \textit{plug-in tau} that achieves \Eqref{eq:good_plug-in_tau} is expected to preserve the difference of the true performance metric and to minimize the upper bound of the finite sample uncertainty term $\mathcal{W}$ in \Eqref{eq:estimator_decomposition}. 

\subsection{Obtaining \textit{plug-in tau}} \label{sec:obtaining_plug-in_tau}
Next, we present a method to obtain a desirable \textit{plug-in tau} inspired by the \textit{doubly robust} (DR) estimator in causal inference and \textit{counterfactual regression} (CFR) in CATE prediction~\cite{bang2005doubly,dudik2011doubly,shalit2017estimating}. 
The proposed procedure is designed to preserve unbiasedness of \textit{plug-in tau} using the DR estimator and to minimize its expected conditional variance with the power of CFR. 
Thus, the idea of combining the DR estimator and CFR is a key to better satisfy \Eqref{eq:good_plug-in_tau}. 
Subsequently, we formally describe the resulting model selection procedure, \textit{counterfactual cross-validation} (CF-CV). 

First, we define a class of \textit{plug-in tau} building on the DR estimator.
\begin{definition}
The doubly robust plug-in tau for a given data $(X, T, Y)$ is defined as follows:
\begin{align}
    & \tiltau_{DR} (\xty; f_t) \notag   \\ 
    & \coloneqq \frac{T-\ps}{\ps(1-\ps)} (Y - f_T(X)) + \oneF - \zeroF , \label{eq:dr_plug-in_tau}
\end{align}
where $f_t: \mathcal{X} \rightarrow \mathcal{Y}$ is an arbitrary regression function.
\end{definition}

We rely on the class of the DR estimator for constructing the \textit{plug-in tau}, because we can design the regression function for a variety of purposes. 
For example, the \textit{more robust doubly robust} estimator utilizes a weighted squared loss to derive the regression function to minimize the variance of the resulting policy value estimator~\cite{farajtabar2018more}. 
In contrast, we can utilize the regression function to minimize the upper bound of the finite sample uncertainty in model selection. 
These objectives cannot be achieved with model-free estimators such as the IPW estimator.

Note that our proposed \textit{plug-in tau} cannot be used for the CATE prediction task because only the feature vectors are available while making predictions. 
In contrast, the treatment assignment and the observed outcome are unavailable. 
Thus, the \textit{plug-in tau} in the form of \Eqref{eq:dr_plug-in_tau} is specialized for the evaluation of CATE predictors.

First, the \textit{plug-in tau} in the form of \Eqref{eq:dr_plug-in_tau} is unbiased against the true CATE as follows.
\begin{proposition} 
Given true propensity scores and a regression function, the proposed plug-in tau is unbiased against the true CATE, i.e.,
\begin{align*}
    \mE\left[\tiltau_{DR} (\xty; f_t)  \givenX \right] = \tau (X).
\end{align*}
See Appendix~\ref{sec:proof_unbiasedness_of_dr_plug-in_tau} for the proof.
\label{prop:unbiasedness_of_dr_plug-in_tau}
\end{proposition}

Next, to consider the condition in \Eqref{eq:good_plug-in_tau}, we state the expected conditional variance of the \textit{plug-in tau}.
\begin{proposition} 
Given true propensity scores and a regression function, the expected conditional variance of the proposed plug-in tau can be represented as:
\begin{align}
& \mE_X \left[ \mV \left(\tiltau_{DR} (\xty; f_t) \givenX  \right) \right] \notag \\
& =  \zeta + \mE_X \left[ \{ \sum_{t\in \mathcal{T}} \sqrt{w_t (X)} (f_t (X) - m_t (X) ) \}^2 \right], \label{eq:variance_of_dr_plug-in_tau}
\end{align}
where 
\begin{align*}
    & w_t (X)  \coloneqq \frac{t(1-2\ps) + \ps^2}{\ps(1-\ps)}, \\
    & \zeta \coloneqq \mE_X \left[ \sum_{t\in \mathcal{T}} \frac{\ps+t(1-2\ps)}{\ps(1-\ps)} (\tY - m_t(X))^2 \right].
\end{align*}
See Appendix~\ref{sec:proof_variance_of_dr_plug-in_tau} for the proof.
\label{prop:variance_of_dr_plug-in_tau}
\end{proposition}

In \Eqref{eq:variance_of_dr_plug-in_tau}, $\zeta$ is independent of $f$.
Thus, we can pursue the minimization of the expected conditional variance of $\tiltau_{DR}$ by training $f$ with the following procedure:
\begin{align}
    \min_{f \in \mathcal{F}} \ \mE_X \left[ \{ \sum_{t\in \mathcal{T}} \sqrt{w_t (X)} (f_t (X) - m_t (X) ) \}^2 \right]
    \label{eq:ideal_loss_for_f},
\end{align}
where $ \mathcal{F} $ is a class of regression functions.
A problem is that the direct minimization of \Eqref{eq:ideal_loss_for_f}  is infeasible, because $ m_0 (x) $ or $ m_1 (x)$ is always counterfactual.
Therefore, we derive the upper bound of the second term of \Eqref{eq:variance_of_dr_plug-in_tau} using only observable variables.
\begin{theorem} 
Let $G$ be a family of functions $g: \mathcal{R} \rightarrow \mathcal{Y}$ and suppose that, for any given $t \in \mathcal{T}$ and $w: \mathcal{X} \times \mathcal{T} \rightarrow \mathcal{R}_{\ge 0}$, there exists a positive constant $\Brep$ such that the per-unit expected loss functions obey $\frac{1}{\Brep} \cdot \tELL (\Psi(r), t) \in G$ where $\Psi$ is the inverse image of $\Phi$. 
Then, the following inequality holds:
\begin{align}
& \mE_X [ \{ \sum_{t\in \mathcal{T}} \sqrt{w_t (X)} (f_t (X) - m_t (X) ) \}^2 ] \notag \\
& \qquad \le 2 \bigl(\epsilon_{F_1}^{w_1} \left(h, \Phi \right) + \epsilon_{F_0}^{w_0}\left(h, \Phi \right) \notag \\
 & \qquad + \Brep \ipm \left( \tPrep, \cftPrep \right) -2 \sigma^2 \bigr), \label{eq:factual_upper_bound_of_variance}
\end{align}
where $\sigma^2 \coloneqq \min_{(t, t^{\prime}) \in \mathcal{T}^2} \{  \sigma_{t, w_t}^2 \left( p_{t^{\prime}} \right)  \} $, and 
\begin{align*}
    & \sigma_{t, w}^2( p_{t^{\prime}} )  \\
    & \coloneqq \int_{\mathcal{X} \times \mathcal{Y}} w(x) (\tY - \tM )^2 p (\tY | x) p_{t^{\prime}} (x) d\tY dx .
\end{align*} 
See Appendix~\ref{sec:proof_factual_upper_bound} for the proof.
\label{thm:factual_upper_bound}
\end{theorem}

\Eqref{eq:factual_upper_bound_of_variance} consists of factual losses and an IPM on the representation space, and thus can be estimated from observed samples. 
The intuition is that the counterfactual losses can be upper bounded by the sum of weighted factual losses and IPM between distributions of the treated and the controlled.
Therefore, we can optimize the upper bound of the expected conditional variance of $\tiltau_{DR}$ using only factual samples in a manner similar to CFR~\cite{shalit2017estimating}.
Thus, we build on the CFR's structure and define our regression function as $f_t (x) = h \left( \Phi(x) , t \right)$.
We then consider the following empirical approximation of \Eqref{eq:factual_upper_bound_of_variance} as a loss to derive a hypothesis $h$ and representation function $\Phi$:
\begin{align}
  h, \Phi 
  & = \min_{h, \Phi} \underbrace{ \sum_{i=1}^{n} \frac{w^{\prime}_t (x_i)}{n} \cdot L\left(h\left(\Phi\left(x_{i}\right), t_{i}\right), y_{i}\right)}_{\textit{empirical weighted risk}} \notag \\
  & + \underbrace{\alpha \ipm \left(\left\{\Phi\left(x_{i}\right)\right\}_{i : t_{i}=0},\left\{\Phi\left(x_{i}\right)\right\}_{i : t_{i}=1}\right)}_{\textit{distributional distance}}. \label{eq:loss_of_f}
\end{align}
where $ w^{\prime}_t (x_i) = \frac{w_t (x_i)}{2} \left( \frac{t_{i}}{\hat{\pi}_1 }+\frac{1-t_{i}}{\hat{\pi}_0} \right) $ and $ \hat{\pi}_t = n^{-1} \sum_{i=1}^{n} \mathbb{I} \{ t_i = t \} $.

We use parameterized deep neural networks for $ \Phi(x) $ and $  h \left( \Phi(x) , t \right) $ and train them in an end-to-end manner using the Adam optimizer~\cite{kingma2014adam}.
$\alpha$ is a trade-off hyperparameter which is a replacement of the incomputable factor $\Brep$. 
We use the \textit{Wasserstein distance}~\cite{shalit2017estimating,cuturi2014fast} as $\ipm$ in the experiments.

The derived \textit{plug-in tau} is unbiased for the true metric and minimizes the upper bound of the controllable term in its expected conditional variance in \Eqref{eq:factual_upper_bound_of_variance}, enabling the accurate and stable model selection of CATE predictors.
Algorithm 1 summarizes the resulting model selection procedure.

\begin{algorithm}[tb]
\caption{Counterfactual Cross-Validation (CF-CV)}
\label{alg1}
\begin{algorithmic}[1]
    \REQUIRE A set of candidate CATE predictors $\mathcal{M} = \{ \hattau_1, ..., \hattau_{|\mathcal{M}|} \} $; an observational validation dataset $ \mathcal{V} = \{ X_i, T_i, Y \}_{i=1}^n $; and a trade-off hyperparameter $\alpha$.
    \STATE Train $f \left( X, T \right)$ by minimizing \Eqref{eq:loss_of_f} using $\mathcal{V}$.
    \STATE Estimate the propensity score (if needed).
    \STATE Calculate the plug-in tau $ \tilde{ \tau }_{DR} $ of samples in $\mathcal{V}$.
    \STATE Estimate performance of candidate predictors in $\mathcal{M}$ based on the performance estimator $\estR$ and $\tilde{ \tau }_{DR}$.
    \ENSURE A selected predictor: $ \hattau^{*} = \argmin_{ \hattau \in \mathcal{M} } \, \estR \left( \hattau \right) $.
\end{algorithmic}
\end{algorithm}

\begin{table*}[ht]
\centering
\caption{Comparison of Model Selection and Hyperparameter Tuning Performance of Alternative Evaluation Metrics.}
\large
\vskip 0.05in
\resizebox{\linewidth}{!}{
\begin{tabular}{cccccccccccccc} 
\toprule
&& \multicolumn{2}{c}{\textbf{Rank Correlation}} && \multicolumn{2}{c}{\textbf{Regret}} && \multicolumn{2}{c}{\textbf{NRMSE}} \\ \cmidrule{3-4}\cmidrule{6-7}\cmidrule{9-10}
\textbf{Methods} && \textbf{\std{Mean}{StdErr}} & \textbf{Worst-Case} && \textbf{\std{Mean}{StdErr}} & \textbf{Worst-Case} && \textbf{\std{Mean}{StdErr}} & \textbf{Worst-Case}  \\ 
\midrule \midrule
IPW && \std{0.195}{0.039} & -0.749  && \std{1.032}{0.100} & 6.779 && \std{0.336}{0.013} & 0.737 \\
$\tau$-risk && \std{0.312}{0.030} & -0.553 && \std{1.392}{0.130} & 7.884 && \std{0.324}{0.013} & 0.700  \\
Plug-in && \std{0.914}{0.006} & 0.591 &  & \std{0.073}{0.012} & 0.780  && \std{0.257}{0.010} & 0.490  \\ 
\midrule
CF-CV (ours) && \best{\std{0.921}{0.005}} & \best{0.666} && \best{\std{0.066}{0.012}} & \best{0.562} && \best{\std{0.256}{0.009}} & \best{0.483}  \\ 
\bottomrule
\end{tabular}}
\vskip 0.05in
\raggedright
\fontsize{9.5pt}{10.5pt}\selectfont \textit{Notes}: 
Mean with standard errors (StdErr), and worst-case performance of the compared evaluation metrics over 100 realizations are reported. 
The \best{red fonts} represent the best performance in each performance measure.
\label{tab:model_selection_results}
\end{table*}

\section{Experiments}
We compare our proposed evaluation procedure and other existing heuristics using a standard semi-synthetic dataset.\footnote{Our code used to conduct the semi-synthetic experiments is available at \href{https://github.com/usaito/counterfactual-cv}{https://github.com/usaito/counterfactual-cv}}

\subsection{Experimental Setup}

We used the Infant Health Development Program (IHDP) dataset provided by~\cite{hill2011bayesian}. 
The original data is obtained from a randomized study of the impact on educational and follow-up interventions on cognitive development of children~\cite{hill2011bayesian,shalit2017estimating,alaa2018limits,yao2018representation}. 
This is a standard semi-synthetic dataset of 747 children with 25 features and has been widely used to evaluate CATE prediction models~\cite{shalit2017estimating,yoon2018ganite,alaa2018limits,yao2018representation,johansson2020generalization}. 
To enable evaluation with the ground-truth CATE, the outcome of this dataset was synthesized by applying several different stochastic models on the observed features.
Moreover, to introduce confounding, a biased subset of the treatment group was removed.
Note that we did not use real-world causal inference datasets such as \textit{jobs} and \textit{twins}~\cite{yoon2018ganite,shalit2017estimating}, because they do not contain the ground-truths for the true CATE and consequently are unable to perform the \textit{evaluation of evaluation metrics}.

We compared the following evaluation metrics in model selection and hyperparameter tuning tasks:

\noindent
\textbf{\underline{(i) IPW validation}}~\cite{gutierrez2017causal,schuler2018comparison}: 
This metric utilizes the following performance estimator:
\begin{align*}
  \hat{\mathcal{R}}_{IPW} (\hattau) = \frac{1}{n} \sum_{i=1}^n \left(  \tiltau_{IPW}(X_i, T_i, Y_i) -  \hattau(X_i)  \right)^2
\end{align*}
where 
$$ \tiltau_{IPW}(X_i, T_i, Y_i) = \frac{T_i}{e(X_i)}Y_i - \frac{1 - T_i}{1 - e(X_i)}Y_i  $$
is used as a \textit{plug-in-tau} that satisfies the unbiasedness for the CATE.

\noindent
\textbf{\underline{(ii) Plug-in validation}}: 
This uses predicted values of potential outcomes by an arbitrary machine learning algorithm as the \textit{plug-in tau} of the performance estimator in \Eqref{eq:performance_estimator}. 
\begin{align*}
    \hat{\mathcal{R}}_{\textit{plug-in}}\ (\hattau) = \frac{1}{n} \sum_{i=1}^n \left(  (\tiltau^{(1)}_i - \tiltau^{(0)}_i) -  \hattau(X_i)  \right)^2
\end{align*}
where $\tiltau^{(1)}_i$ and $\tiltau^{(0)}_i$ are predictions for the potential outcomes. 
We used CFR~\cite{shalit2017estimating} to construct $\tiltau^{(1)} (\cdot)$ and $\tiltau^{(0)} (\cdot)$ to ensure a fair comparison.

\textbf{\underline{(iii) $\tau$-risk}}~\cite{schuler2018comparison}: 
This metric is derived from the loss function of R-learner in~\cite{nie2017quasi} and is defined as follows:
\begin{align*}
     \hat{\mathcal{R}}_{\tau} (\hattau) =\frac{1}{n} \sum_{i=1}^n \left((Y_i - m (X_i)) - (T_i - e(X_i)) \hattau(X_i)  \right)^2 
\end{align*}
where $m (\cdot) $ is the expectation of observed outcome $\mE [\obsY | X] $. 
We used gradient boosting regressor (GBR) implemented in \textit{scikit-learn} to estimate this parameter.

\textbf{\underline{(iv) Counterfactual Cross-Validation}}: 
This is our proposed metric, which relies on the \textit{plug-in tau} in \Eqref{eq:dr_plug-in_tau}. 
The hyperparameter tuning procedure to derive the regression function $f$ can be found in Appendix~\ref{sec:experimental_settings_model_selection}.

We used logistic regression to estimate the propensity score for CF-CV and IPW validation, because the true propensity score is generally unknown in real-world situations. For plug-in validation and CF-CV, we used the $\mu$-risk~\cite{schuler2018comparison} as a data-driven heuristic to tune hyperparameters of machine learning models to obtain predictions of the potential outcomes or the regression function.

\subsection{Model Selection Performance}
We first tested the model selection performance. \\

\noindent
\textbf{Experimental Procedure.}
We followed the experimental procedure in~\cite{schuler2018comparison}; 
We trained candidate predictors on the training set and made predictions on both validation and test sets. 
Then, we ranked those predictors based on each evaluation metric on the validation set. 
Finally, we compare these estimated performances on the validation set and the true performance on the testing set. 
We conducted the experimental procedure over 100 different realizations with 35/35/30 train/validation/test splits. \\

\noindent
\textbf{Candidate Models.}
We constructed a set of candidate predictors $\mathcal{M}$ by combining five machine learning algorithms (decision tree, random forest, gradient boosting tree, ridge regressor, and support vector regressor) implemented in \textit{scikit-learn} and five meta-learners (S-learner, X-learner, T-learner, domain adaptation learner, and doubly robust learner) implemented in \textit{EconML}\footnote{https://econml.azurewebsites.net/}. 
Thus, we had a set of 25 CATE predictors to select among (i.e., $|\mathcal{M}| = 25$). \\

\begin{figure*}[ht]
\centering
\begin{center}
    \begin{tabular}{c}
        \begin{minipage}{0.49\hsize}
            \begin{center}
                \includegraphics[clip, width=8.5cm]{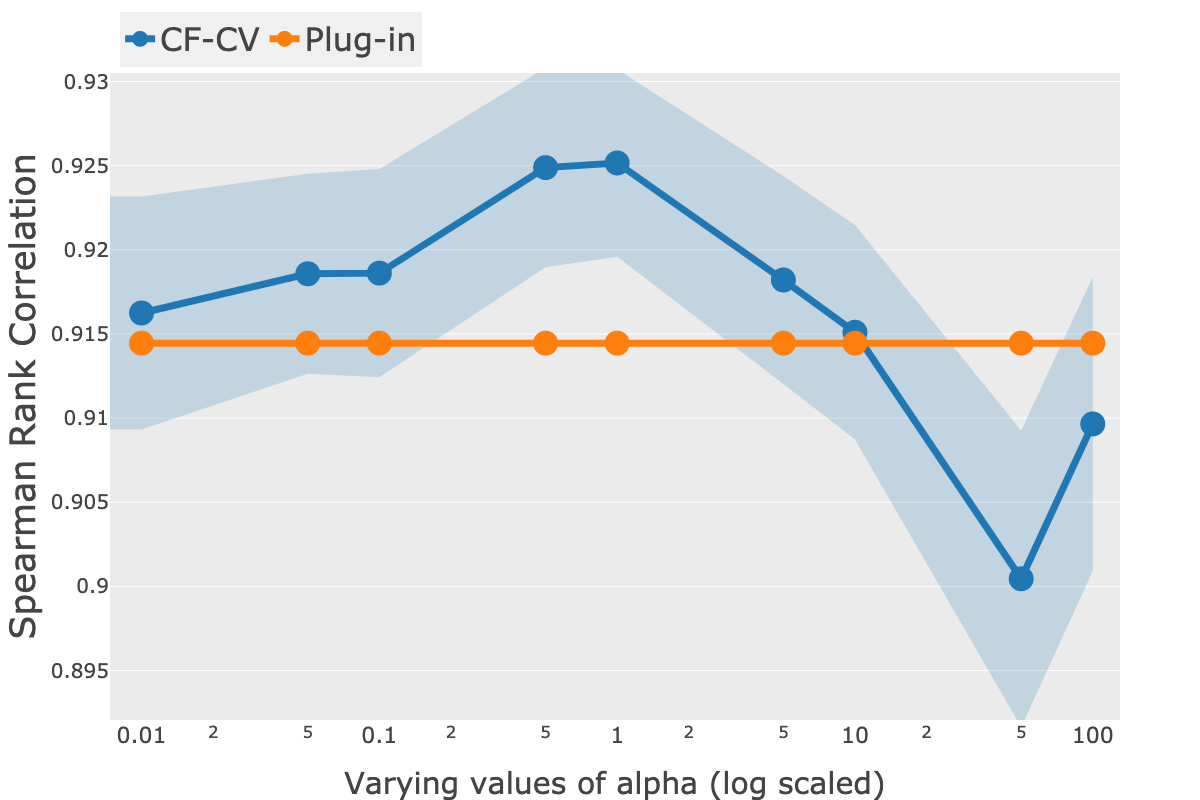}
                \vskip 0.05in
                (a) \textit{Rank correlation} of CF-CV with different values of $\alpha$ 
            \end{center}
        \end{minipage}
        
        \begin{minipage}{0.49\hsize}
            \begin{center}
                \includegraphics[clip, width=8.5cm]{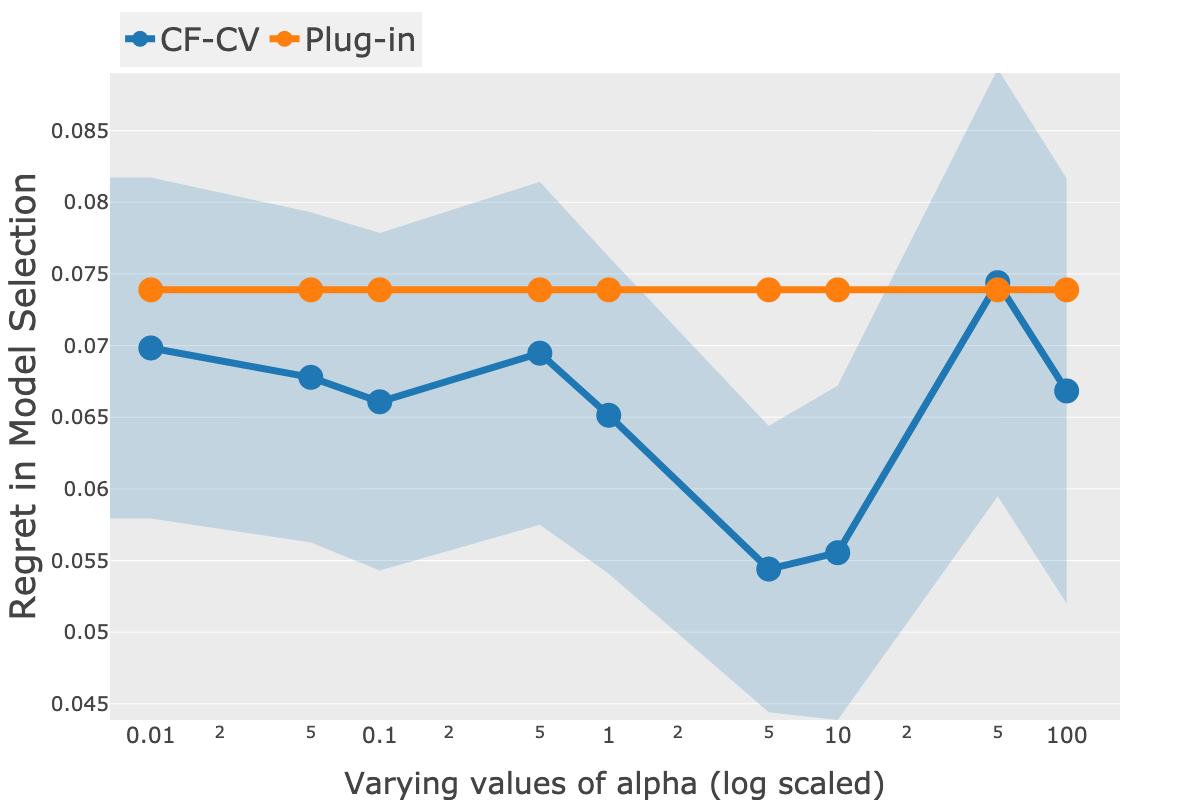}
                \vskip 0.05 in
                (b) \textit{Regret} of CF-CV with different values of $\alpha$ 
            \end{center}
        \end{minipage}
        
    \end{tabular}
\end{center}
\caption{
Comparing CF-CV with varying $\alpha$ and the plug-in validation. 
CF-CV (the blue lines) outperforms the plug-in validation (the orange lines) in most cases and demonstrates its robustness to the choice of $\alpha$.}
\end{figure*}

\noindent
\textbf{Results.}
\Tabref{tab:model_selection_results} reports the mean and worst-case performances over 100 realizations. 
We evaluated the worst-case model selection performance, because we never know the ground-truth performance of any predictor in the real-world, and stable model selection performance is essential. 
\textit{Rank correlation} is the Spearman rank correlation between the rankings by the true performance and the estimated metric values. \textit{Regret in model selection} is the difference between the true performance of the selected model and that of the best possible candidate in $\mathcal{M}$, which is defined as:
\begin{align*}
    \textit{Regret}  = \frac{\trueR \left( \hattau_{selected} \right) - \trueR \left( \hattau_{best} \right) }{ \trueR \left( \hattau_{best} \right) }
\end{align*}
where $ \hattau_{selected} = \argmin_{ \hattau \in \mathcal{M} } \estR (\hattau) $ is the model selected by $\estR$ and $ \hattau_{best} = \argmin_{ \hattau \in \mathcal{M} } \trueR \left( \hattau \right) $ is the best model in $\mathcal{M}$.

\Tabref{tab:model_selection_results} shows the effective model selection performance of the proposed CF-CV. 
In particular, it significantly outperformed the others in terms of the worst-case performance. 
This result empirically suggests that the proposed metric can stably select a well-performing CATE predictor among potential candidates and is an appropriate choice for real-world situations. 
The stability of CF-CV could be a result of its variance upper bound minimization property.
The improvement of the worst-case performance is essential in many causal inference problems such as personalized medicine, which has a great impact on human lives.
Our procedure thus helps avoid deploying poor-performing CATE predictors and enable the safe uses of causal inference in practice.
We also evaluated the sensitivity of the proposed metric to changes in the trade-off hyperparameter $\alpha$.
Figure 1 shows the performances of CF-CV with variation of $\alpha$ compared to the performance of the plug-in validation. 
For the \textit{rank correlation}, CF-CV generally outperformed the plug-in metric with small values of $\alpha$, although it was slightly outperformed by the plug-in with a larger $\alpha$. 
Additionally, CF-CV consistently outperformed the plug-in validation with all values of $\alpha$ in \textit{regret}. 
These results suggest that the proposed metric is robust to the choice of $\alpha$.


\subsection{Hyperparameter Tuning Performance}
Next, we compared the hyperparameter tuning performance. \\

\noindent
\textbf{Tuned Model.}
We tuned the hyperparameters of the combination of GBR and domain adaptation learner (DAL) implemented in \textit{scikit-learn} and \textit{EconML}, respectively. 
DAL consists of three base learners including \textbf{treated\_model}, \textbf{controls\_model}, and \textbf{overall\_model}. Thus, we aimed to find the best three sets of hyperparameters of GBR to optimize the resulting CATE prediction model. \\

\noindent
\textbf{Experimental Procedure.}
We used \textit{Optuna}~\cite{akiba2019optuna} to tune the CATE predictor and set each metric as its objective function. 
For each metric, we sought 100 points in the hyperparameter search space\footnote{The hyperparameter search space is described in Appendix~\ref{sec:experimental_settings_hyperparameter_tuning}}. 
The hyperparameter tuning performance of each metric was evaluated by the true performance of the tuned model on the testing set. 
We repeated the experimental procedure with 100 different realizations and train/validation/test splits. \\

\noindent
\textbf{Results.}
\Tabref{tab:model_selection_results} provides the results of the hyperparameter tuning experiment.
We report the mean and worst-case \textit{normalized root-mean-squared-error} (NRMSE) of CATE predictors tuned by each metric defined below\footnote{We used NRMSE, as the potential outcomes of the IHDP dataset have different scales among realizations.}.
\begin{align*}
    \textit{NRMSE} = \sqrt{ \frac{ n^{-1} \sum_{i=1}^n ( \tau (X_i) - \hattau (X_i) )^2  }{ \hat{ \mV } ( \tau (X) ) } }
\end{align*}
where $\{ \hattau (X_i) \}_{i=1}^n$ is a set of CATE predictions by $ \hattau (\cdot) $ and $ \hat{ \mV } ( \tau )  $ is an empirical variance of the ground-truth CATE.

\Tabref{tab:model_selection_results} shows that our metric improved the worst-case performance by 1.4 \% compared to the best baselines. 
Although the mean NRMSE is almost the same as that with the plug-in validation, the results demonstrate that the proposed metric allows stable hyperparameter tuning of the CATE prediction models.
\section{Conclusion}
In this work, we studied the model selection problem in CATE prediction. 
In contrast to previous studies, we aimed to identify the rank order of the true prediction performances of the candidate prediction models. 
We achieved this by using a modified version of the CFR as a regression function of the DR estimator to minimize the finite sample uncertainty. 
Empirical evaluations demonstrated the effectiveness and stability of the proposed metric for model selection and hyperparameter tuning of the CATE predictors.

Important future research directions include consideration of situations with hidden confounders, and a possible extension to the \textit{off-policy evaluation} of bandit policies.

\bibliography{cfcv.bbl}
\bibliographystyle{icml2020}

\clearpage
\onecolumn
\icmltitle{Counterfactual Cross-Validation: \\
Stable Model Selection Procedure for Causal Inference Models-- Appendix}

\appendix

\section{Omitted Proofs}

In this section, we denote $\tau(X)$, $\hattau(X) $, and $\tiltau(\xty)$ as $\tau$, $\hattau$, and $\tiltau$ for simplicity.
We also denote $\tau(X_i)$, $\hattau(X_i) $, and $\tiltau(X_i, T_i, Y_i)$ as $\tau_i$, $\hattau_i$, and $\tiltau_i$.

\subsection{Proof of \Propref{prop:rank_identification}}
\label{sec:proof_rank_identification}

\begin{proof}
First, the following equality holds:
\begin{align*}
\mE \left[ \estR (\hattau) \right] 
 =  \mE \left[ \frac{1}{n} \sum_{i=1}^{n} (\tiltau_i - \hattau_i)^2  \right] 
 =  \frac{1}{n}\sum_{i=1}^{n}  \mE \left[   (\tiltau - \tau + \tau - \hattau)^2  \right] 
 =  \mE \left[   (\tiltau - \tau )^2  \right] 
  - \frac{2}{n}\sum_{i=1}^{n}  \underbrace{\mE \left[   (\hattau - \tau )(\tiltau - \tau)  \right]}_{(a)} 
  + \underbrace{\mE \left[   (\hattau - \tau )^2  \right]}_{ \trueR (\hattau) }
\end{align*}
Then, we have 
$ (a)  =  \mE \left[   (\hattau - \tau )(\tiltau - \tau)  \right] 
 = \mE \left[ \mE \left[  (\hattau - \tau)( \tiltau - \tau) \ |\ X \right]  \right] 
 = \mE \left[   (\hattau - \tau )  (\mE [ \tiltau \givenX] - \tau )   \right] 
 = 0  
$.

Thus, we obtain $\mE [  \estR \left( \hattau \right) ] = \trueR \left( \hattau \right) + \mE [ \left(  \tiltau - \tau \right)^2 ] $.
\end{proof}

\subsection{Derivation of \Eqref{eq:estimator_decomposition}}
\begin{proof}
Following the same procedure as in the proof of Proposition 1, we have
\begin{align*}
    \estR (\hattau)
     =  \frac{1}{n} \sum_{i=1}^{n} (\tiltau_i - \hattau_i)^2  
      =  \frac{1}{n}\sum_{i=1}^{n}   (\tiltau_i - \tau_i + \tau_i - \hattau_i)^2 
      =  \frac{1}{n} \sum_{i=1}^{n}   (\tiltau_i - \tau_i )^2
      - \frac{2}{n}\sum_{i=1}^{n}   (\tiltau_i - \tau_i ) (\hattau_i - \tau_i )
      + \frac{1}{n} \sum_{i=1}^{n}    (\hattau_i - \tau_i )^2  
  \end{align*}
\end{proof}

\subsection{Proof of \Propref{prop:unbiasedness_of_dr_plug-in_tau}}
\label{sec:proof_unbiasedness_of_dr_plug-in_tau}

\begin{proof}
We rewrite the DR plug-in tau in \Eqref{eq:dr_plug-in_tau} as:
\begin{align*}
\tiltau_{DR}(X, T, Y) & = \frac{T}{\ps} \left( Y - \oneF \right) - \frac{1-T}{1-\ps} \left( Y - \zeroF \right)  + \left( \oneF - \zeroF \right)   \\
& = \tiltau_{DR_1}(\xty) - \tiltau_{DR_0}(\xty) 
\end{align*}
where 
$ \tiltau_{DR_1}(\xty)  = \frac{T}{\ps} \left( Y - \oneF \right) + \oneF $ and
$ \tiltau_{DR_0}(\xty)  = \frac{1-T}{1-\ps} \left( Y - \zeroF \right) + \zeroF  $.
    
Then, the expectation of $\tiltau_{DR_1}$ is 
$ \mE \left[ \tiltau_{DR_1} \givenX \right] 
   = \mE \left[ \frac{T}{\ps}  \givenX \right] \mE \left[ \left( \oneY - \oneF \right)  \givenX \right] + \oneF
   = \mE \left[  \oneY   \givenX \right]  
$.

We also have $ \mE \left[ \tiltau_{DR_0} \givenX \right] = \mE \left[  \zeroY   \givenX \right]$ in the same way. 
Thus, we have, $\mE \left[ \tiltau_{DR} \givenX \right] =  \mE \left[  \oneY  -  \zeroY \givenX \right] = \tau $.
\end{proof}

\subsection{Proof of \Propref{prop:variance_of_dr_plug-in_tau}}
\label{sec:proof_variance_of_dr_plug-in_tau}

\begin{proof}
The second moment of $\tiltau_{DR_1}$ is 
\begin{align*}
 \mE \left[ \left( \tiltau_{DR_1} \right)^2 \givenX \right]  
  & = \mE \left[ \left( \frac{T}{\ps} \left( \oneY - \oneF \right) + \oneF \right)^2 \givenX \right]  \\
  & = \mE \left[ \left( \left(1 - \frac{T}{\ps} \right) \left( \oneF - \oneY  \right) + \oneY \right)^2 \givenX \right]  \\
  & =  \mE \left[ \zeta_1  \givenX \right]  + \left( \oneM \right)^2  + \oneW \left( \oneF - \oneM  \right)^2 
\end{align*}
We also have the second moment of $\tiltau_{DR_0}$ in the same manner as follows:
\begin{align*}
  \mE \left[ \left( \tiltau_{DR_0} \right)^2 \givenX \right] 
  =  \mE \left[ \zeta_0  \givenX \right]  + \left( \zeroM \right)^2 + \zeroW \left( \zeroF - \zeroM  \right)^2 
\end{align*}
where $ \zeta_1 =  (\oneY - \oneM )^2 / \ps $, $ \zeta_0 =  (\zeroY - \zeroM )^2 / (1 - \ps) $.
Note that $ \mE \left[ \zeta_1  \givenX \right] = \mV ( \oneY \givenX ) / \ps $ and $ \mE \left[ \zeta_0  \givenX \right] = \mV ( \zeroY \givenX ) / (1-\ps) $.

Then, by using the result of Proposition 3, we obtain
\begin{align*}
  \mV \left( \tiltau_{DR_1}  \givenX \right) & =  \mE \left[ \zeta_1 \givenX \right]   + \oneW \left( \oneF-\oneM  \right)^2  \\
  \mV \left( \tiltau_{DR_0}  \givenX \right) & =  \mE \left[ \zeta_0 \givenX \right]   + \zeroW \left( \zeroF-\zeroM  \right)^2 
\end{align*}
In addition, from Lemma~\ref{lem:conditional_covariance},
\begin{align*}
  \mV \left( \tiltau_{DR}  \givenX  \right)
  & = \mV \left( \tiltau_{DR_1} - \tiltau_{DR_0}  \givenX \right)  \\
  & = \mV \left( \tiltau_{DR_1}  \givenX \right) - 2\mathrm{Cov} \left( \tiltau_{DR_1}, \tiltau_{DR_0}  \givenX \right)  + \mV \left( \tiltau_{DR_0} \givenX \right)  \\
  & = \mE \left[ \zeta_1 + \zeta_0  \givenX \right]  +  \oneW \left( \oneF - \oneM  \right)^2  \\
  & \quad + \zeroW \left( \zeroF - \zeroM  \right)^2 +2 \left( \oneF - \oneM  \right) \left( \zeroF - \zeroM  \right)  \\
  & = \mE \left[ \zeta_1  + \zeta_0  \givenX \right] + \left( \sqrt{\oneW}\left( \oneF - \oneM  \right) +  \sqrt{\zeroW}\left( \zeroF - \zeroM  \right) \right)^2
\end{align*}
where $ \oneW \zeroW  =1 $.
Hence, we have 
$ \mE_X [ \mV \left( \tiltau_{DR}  \givenX  \right) ]  
= \zeta + \mE_X \left[  \left\{ \sum_{t\in \mathcal{T}} \sqrt{w_t (X)} (f_t (X) - m_t (X) ) \right\}^2 \right]
$ where $ \zeta =  \mE \left[ \zeta_1  + \zeta_0 \right] $ . 
\end{proof}

\subsection{Proof of \Thmref{thm:variance_upper_bound}}
\label{sec:proof_variance_upper_bound}

\begin{proof}
\begin{align*}
    & \mV \left( 2n^{-1} \ \sum_{i=1}^{n}  (\hattau_i - \tau_i )(\tiltau_i - \tau_i )   \right)  \\
    & =  4n^{-2} \ \mV \left( \sum_{i=1}^{n}  (\hattau_i - \tau_i )(\tiltau_i - \tau_i ) \right) \\
    & = 4n^{-2} \ \mE \left[ \left(  \sum_{i=1}^{n}  (\hattau_i - \tau_i )(\tiltau_i - \tau_i ) \right)^2  \right]  \quad \because (a) = 0  \\
    & = 4n^{-2} \ \mE \left[  \sum_{i=1}^{n} \sum_{j=1}^{n} (\hattau_i - \tau_i )(\tiltau_i - \tau_i ) (\hattau_j - \tau_j )(\tiltau_j - \tau_j )  \right] \\
    & = 4n^{-2} \ \sum_{i=1}^{n} \mE \left[  (\hattau_i - \tau_i )^2 (\tiltau_i - \tau_i )^2 \right] 
    \quad \because \mE_X [ (\hattau_i - \tau_i )(\tiltau_i - \tau_i )(\hattau_j- \tau_j )(\tiltau_j - \tau_j ) ] = 0, \forall i, j (i\neq j)  \\
    & \le  4 \mC n^{-1} \ \mE \left[ (\tiltau - \tau )^2  \right] \\
    & = 4 \mC n^{-1} \ \mE_X \left[ \mE [ (\tiltau - \mE [ \tiltau \givenX ] )^2 ] \givenX \right] \\
    & =  4 \mC n^{-1} \  \mE_X \left[ \mV \left(\tiltau \givenX  \right) \right]
  \end{align*}
\end{proof}

\subsection{Technical Lemmas}

\begin{lemma}
The conditional covariance of $\tiltau_{DR_1}$ and $\tiltau_{DR_0}$ is:
\begin{align*}
  \mathrm{Cov} \left( \tiltau_{DR_1}, \tiltau_{DR_0}  \givenX  \right) = - \left( \oneF - \oneM  \right) \left( \zeroF - \zeroM  \right)
\end{align*}
\begin{proof}
  \begin{align*}
    \mathrm{Cov} \left( \tiltau_{DR_1}, \tiltau_{DR_0}  \givenX  \right) 
    & = \mE \left[ \tiltau_{DR_1} \cdot \tiltau_{DR_0}  \givenX  \right] -  \mE \left[ \tiltau_{DR_1} \givenX \right] \cdot \mE \left[ \tiltau_{DR_0} \givenX \right]   \\
    & = \mE \left[ \tiltau_{DR_1} \cdot \tiltau_{DR_0}  \givenX \right] - \oneM  \cdot \zeroM 
  \end{align*}
  Then,
\begin{align*}
  & \mE \left[ \tiltau_{DR_1} \cdot \tiltau_{DR_0}  \givenX \right]  \\
  & = \oneF\zeroF  + \oneF \mE \left[  \frac{1-T}{1-\ps} \left( \zeroY - \zeroF \right) \givenX \right] +  \zeroF \mE \left[  \frac{T}{\ps} \left( \oneY - \oneF \right) \ | \   X \right]  \\
  & = \oneF\zeroF + \oneF(\zeroM - \zeroF) + \zeroF(\oneM - \oneF) 
\end{align*}
Therefore,
\begin{align*}
  & \mE \left[ \tiltau_{DR_1} \cdot \tiltau_{DR_0}  \givenX \right] - \oneM \zeroM \\
  & =  \oneF\zeroF + \oneF(\zeroM - \zeroF) + \zeroF(\oneM - \oneF) - \oneM \zeroM  \\
  & = - \left( \oneF - \oneM  \right) \left( \zeroF - \zeroM  \right) 
\end{align*}
\end{proof}
\label{lem:conditional_covariance}
\end{lemma}

\begin{lemma}(Similar to Lemma A.4 of~\cite{shalit2017estimating})
Let $\Phi: \mathcal{X} \rightarrow \mathcal{R} $ be an invertible representation with $ \Psi $ its inverse.
Let $G$ be a family of functions $g : \mathcal{R} \rightarrow \mathbb{R}_{\ge 0}$ 
and $h : \mathcal{R} \times \mathcal{T} \rightarrow \mathcal{Y}$ be a hypothesis. 
Assume that, for any given $t \in \mathcal{T}$ and $w: \mathcal{X} \times \mathcal{T} \rightarrow \mathcal{R}_{\ge 0}$, there exists a constant $\Brep>0$, such that $\frac{1}{\Brep} \cdot \tELL(\Psi(r), t) \in \mathrm{G}$. 
Then we have:
\begin{align*}
  \epsilon_{CF_{1-t}}^w (h, \Phi)  \le \ftEPS  + \Brep \cdot \ipm \left( \tPrep,  \cftPrep \right)
\end{align*}
\begin{proof}
  \begin{align*}
    \epsilon_{CF_{1-t}}^w (h, \Phi) - \ftEPS
    & = \int_{\mathcal{X}} \tELL (x, t) \left( \cftP -  \tP\right) dx  \\
    & = \int_{\mathcal{R}} \tELL (\Psi(r) , t) \left( \cftPrep -  \tPrep\right) dr \quad \because \text{Lemma A.2 of~\cite{shalit2017estimating}} \\
    & = \Brep \cdot \int_{\mathcal{R}} \frac{ \tELL (\Psi(r) , t)}{\Brep } \left( \cftPrep -  \tPrep \right) dr  \\
    & \le \Brep \cdot \sup_{g \in G} \left| \int_{\mathcal{R}} g(r) \left( \cftPrep -  \tPrep \right) dr \right|   \\
    & =  \Brep \cdot \ipm \left(\tPrep,  \cftPrep \right)
  \end{align*}
where Lemma A.2 of~\cite{shalit2017estimating} states the standerd changes of variable formula: $  p^{\Phi} (t \ | \ r) = p(t \ | \ \Psi(r)) $ and $p^{\Phi} (\tY \ | \ r) = p(\tY \ | \ \Psi(r)) $ for all $r \in \mathcal{R}$ and $t \in \mathcal{T}$.
\end{proof}
\label{lem:first_lemma}
\end{lemma}

\begin{lemma}(Similar to Lemma A.5 of~\cite{shalit2017estimating})
Let $\Phi: \mathcal{X} \rightarrow \mathcal{R} $ be an invertible representation and $h : \mathcal{R} \times \mathcal{T} \rightarrow \mathcal{Y}$ be a hypothesis.
We also define a regression function as $f_t (x) = h (\Phi (x), t ) $.
Then, for any given $t \in \mathcal{T}$ and $w: \mathcal{X} \times \mathcal{T} \rightarrow \mathcal{R}_{\ge 0}$, the following equalities hold:
\begin{align*}
  \int_{\mathcal{X} } w(x)  \left( \tF - \tM \right)^2  \tP  dx  
  & = \ftEPS - \sigma_{t, w}^2 (p_t)  \\
  \int_{\mathcal{X} } w(x)  \left( \tF - \tM \right)^2  \cftP  dx  
  & = \epsilon_{CF_{1-t}}^w (h, \Phi) - \sigma_{t, w}^2 ( p_{1-t})
\end{align*}
\begin{proof}
  \begin{align*}
    \ftEPS 
    & = \int_{\mathcal{X}} \tELL (x, t)  \tP  dx \\
    & =  \int_{\mathcal{X} \times \mathcal{Y}} w(x) \left( \tF - \tY \right)^2 p \left(\tY | x \right) \tP d\tY dx  \\
    & = \int_{\mathcal{X}} w(x)  \left( \tF - \tM \right)^2  \tP  dx \\
    & \quad - 2 \int_{\mathcal{X} \times \mathcal{Y}} w(x)  \left( \tF - \tM \right) \left( \tY - \tM \right)  p \left(\tY, x  \ | \ t \right) d\tY dx  \\
    & \quad + \int_{\mathcal{X} \times \mathcal{Y}} w(x) \left( \tY - \tM  \right)^2 p \left(\tY , x \ | \ t \right) d\tY dx  \\
     & = \int_{\mathcal{X}} w(x)  \left( \tF - \tM \right)^2  \tP dx + \sigma_{t, w}^2 (p_t) 
  \end{align*}
  Thus, we have,
  \begin{align*}
    \int_{\mathcal{X} } w(x)  \left( \tF - \tM \right)^2  \tP  dx = \ftEPS - \sigma_{t, w}^2 (p_t)  
  \end{align*}
  We can derive the analogous equality for counterfactual loss in the same manner.
\end{proof}
\label{lem:second_lemma}
\end{lemma}

\subsection{Proof of \Thmref{thm:factual_upper_bound}}
\label{sec:proof_factual_upper_bound}

\begin{proof}
\begin{align*}
&  \mE_X [ \{ \sum_{t\in \mathcal{T}} \sqrt{w_t (X)} (f_t (X) - m_t (X) ) \}^2 ] \\
    &  = \mE_X [ \{ \sqrt{ \oneW } (\oneF - \oneM ) + \sqrt{ \zeroW } (\zeroF - \zeroM ) \}^2 ] \\
    & \le 2 \int_{\mathcal{X}} \left( \oneW \left( \oneF - \oneM  \right)^2 
    +  \zeroW \left( \zeroF - \zeroM  \right)^2 \right) p(x) \ dx \quad \because (x + y)^2 \le 2(x^2 + y^2)  \\
    & = 
    2\pi_1 \int_{\mathcal{X}}  \oneW \left( \oneF - \oneM  \right)^2  \oneP \ dx  
    + 2\pi_0 \int_{\mathcal{X}}  \oneW \left( \oneF - \oneM  \right)^2  \zeroP \ dx  \\
    & \quad + 2\pi_1 \int_{\mathcal{X}}  \zeroW \left( \zeroF - \zeroM  \right)^2  \oneP \ dx  
     + 2\pi_0 \int_{\mathcal{X}}  \zeroW \left( \zeroF - \zeroM  \right)^2  \zeroP \ dx  \\
    & = 
    2\pi_1 \left(\epsilon^{w_1}_{F_1} (h, \Phi) - \sigma_{t=1, w_1}^2 (p_1) \right) 
    + 2\pi_0 \left(\epsilon^{w_1}_{CF_0} (h, \Phi) - \sigma_{t=1, w_1}^2 (p_0) \right)  \\
    & \quad 
    + 2\pi_1 \left(\epsilon^{w_0}_{CF_1} (h, \Phi) - \sigma_{t=0, w_0}^2 (p_1) \right)  
    +  2\pi_0 \left(\epsilon^{w_0}_{F_0} (h, \Phi) - \sigma_{t=0, w_0}^2 (p_0) \right) 
    \quad \because \text{\Lemref{lem:second_lemma}}  \\
    & \le 2\epsilon^{w_1}_{F_1} (h, \Phi) +  2\epsilon^{w_0}_{F_0} (h, \Phi) + 2 \Brep \cdot \ipm \left(\tPrep,  \, \cftPrep \right)  - 4\sigma^2  \quad \because \text{\Lemref{lem:first_lemma}}
\end{align*}
\end{proof}

\section{Detailed Experimental Settings}
\label{sec:experimental_settings}

\subsection{Model Selection Experiment in Section 5.2}
\label{sec:experimental_settings_model_selection}

The weighted counterfactual regression model used as a regression function of our proposed metric has some hyperparameters itself. To tune the hyperparameters of this model, we used the simple $\mu$-risk as described in~\cite{schuler2018comparison}. 
\Tabref{tab:search_space_for_f} describes the hyperparameter search spaces and the resulting set of hyperparameters for the weighted counterfactual regression .

\begin{table}[h]
\caption{Hyperparameter search spaces and the selected values of the hyperparameters for the weighted counterfactual regression in our proposed CFCV. A set of hyperparameters optimzed the $\mu$-risk~\cite{schuler2018comparison} was selected for the weighted CFR.}
\centering
\vskip 0.1in
\begin{tabular}{l|cc} 
\toprule
    Hyperparameters  & Search spaces & Selected values  \\ 
\midrule
    Num. of hidden layers for $h$ and $\Phi$ in \Eqref{eq:loss_of_f} & $ \{1, 2, 3\} $ & 3 \\
    Dim. of hidden layers for $h$ and $\Phi$ in \Eqref{eq:loss_of_f} &  $\{20, 50, 100\}$ & 100 \\
    trade-off parameter $\alpha$ in \Eqref{eq:loss_of_f} & $[0.01, 100]$ & 0.356 \\ 
    learning\_rate & $[0.0001, 0.01]$ & $4.292 \times 10^{-4} $ \\
    batch\_size & 256 (fixed) & 256 (fixed)  \\
    dropout rate & 0.2 (fixed) & 0.2 (fixed) \\ 
\bottomrule
\end{tabular}
\label{tab:search_space_for_f}
\end{table}

\subsection{Hyperparameter Tuning Experiment in Section 5.3}
\label{sec:experimental_settings_hyperparameter_tuning}

\Tabref{tab:search_space_for_gbr} provides the hyperparameter search space of the Gradient Boosting Regressor used in the hyperparameter tuning experiment in Section 5.3.

\begin{table}[h]
\centering
\caption{Hyperparameter search space for Gradient Boosting Regressors.}
\vskip 0.1in
\begin{tabular}{l|c} 
\toprule
    Hyperparameters  & Search spaces  \\ 
\midrule
    n\_estimators & $100$ (fixed)  \\
    max\_depth &  $[1, 20]$  \\
    min\_samples\_leaf  & $[1, 20]$  \\ 
    learning\_rate & $[10^{-5}, 10^{-1}]$ \\
    subsample & $\{0.1, 0.2, \ldots 1.0\}$  \\ 
\bottomrule
\end{tabular}
\label{tab:search_space_for_gbr}
\end{table}

\end{document}